\documentclass[10pt,twocolumn,letterpaper]{article}
\usepackage{multirow}
\usepackage{makecell}
\usepackage{caption}

\usepackage{adjustbox}
\usepackage{wrapfig}

\usepackage{float}
\usepackage{subfig}

\usepackage{color, colortbl}
\usepackage{cvpr}              
\usepackage{marvosym}
\usepackage[accsupp]{axessibility}
\definecolor{cvprblue}{rgb}{0.21,0.49,0.74}
\usepackage[pagebackref,breaklinks,colorlinks,allcolors=cvprblue]{hyperref}

\def\paperID{9944} 
\def\confName{CVPR}
\def\confYear{2025}
\makeatletter
\def\thanks#1{\protected@xdef\@thanks{\@thanks
        \protect\footnotetext{\hspace{-2em}#1}}}
\makeatother

\title{Adaptive Markup Language Generation for Contextually-Grounded Visual Document Understanding}

\author{
Han Xiao$^{1,2\dagger}$, Yina Xie$^{2}$, Guanxin Tan$^{2}$, Yinghao Chen$^{2}$, Rui Hu$^{2}$, Ke Wang$^{1}$,\\ Aojun Zhou$^{1}$, Hao Li$^{1}$, Hao Shao$^{1}$, Xudong Lu$^{1,2\dagger}$,  
Peng Gao$^{4}$,\\ Yafei Wen$^{2}$, Xiaoxin Chen$^{2}$, Shuai Ren$^{2\,\ddagger\,{\textrm{\Letter}}}$, Hongsheng Li$^{1,3\,{\textrm{\Letter}}}$\\
\text{$^1$CUHK MMLab\quad $^2$vivo AI Lab\quad $^3$CPII under InnoHK}\\
\text{\quad$^4$Shanghai AI Lab \& Shenzhen Institute of Advanced Technology, CAS}\\
\texttt{\{1155229123@link,hsli@ee\}.cuhk.edu.hk}\\
\texttt{shuai.ren@vivo.com}\thanks{$^{\textrm{\Letter}}$Corresponding author $^\ddagger$Project lead $^\dagger$Interns at vivo.\\}
}

\begin{document}
\maketitle
\begin{abstract}
Visual Document Understanding has become essential with the increase of text-rich visual content. This field poses significant challenges due to the need for effective integration of visual perception and textual comprehension, particularly across diverse document types with complex layouts. Moreover, existing fine-tuning datasets for this domain often fall short in providing the detailed contextual information for robust understanding, leading to hallucinations and limited comprehension of spatial relationships among visual elements. To address these challenges, we propose an innovative pipeline that utilizes adaptive generation of markup languages, such as Markdown, JSON, HTML, and TiKZ, to build highly structured document representations and deliver contextually-grounded responses. 
We introduce two fine-grained structured datasets: DocMark-Pile, comprising approximately 3.8M pretraining data pairs for document parsing, and DocMark-Instruct, featuring 624k fine-tuning data annotations for grounded instruction following.
Extensive experiments demonstrate that our proposed model significantly outperforms existing state-of-the-art MLLMs across a range of visual document understanding benchmarks, facilitating advanced reasoning and comprehension capabilities in complex visual scenarios.
Our code and models are released at \url{https://github.com/Euphoria16/DocMark}.
\end{abstract}    
\section{Introduction}
\label{sec:intro}

Text-rich visual content, including multi-page documents and various multimedia formats, is ubiquitous in our daily lives. Consequently, Visual Document Understanding has emerged as a fundamental task that encompasses document classification, information extraction, and document-related visual question answering. 
This area presents significant challenges, as it requires seamless integration of visual perception with textual comprehension, along with the ability to navigate the complex interactions between these 
modalities.
Such capabilities not only represent a fundamental cognitive skill in humans, but are also essential for the advancement of artificial intelligence systems. 

In recent years, Multimodal Large Language Models (MLLMs) have demonstrated remarkable progress in handling multi-modal tasks.
Notably, there has been a growing focus on enhancing the capabilities of MLLMs in visual document comprehension.
Existing efforts mainly aim to improve the representation of vision encoders originally designed for natural images, which are usually trained on relatively low-resolution datasets. Such efforts include scaling up the input resolution~\cite{liu2024llava, chen2024far, hong2024cogagent, li2024monkey}, retraining vision encoders on text-rich images~\cite{chen2024internvl, wei2025vary, liu2024textmonkey}, and leveraging compositions of different encoders~\cite{lin2023sphinx, li2024mini}.

\begin{figure}[t]
    \centering
    \includegraphics[width=0.98\linewidth]{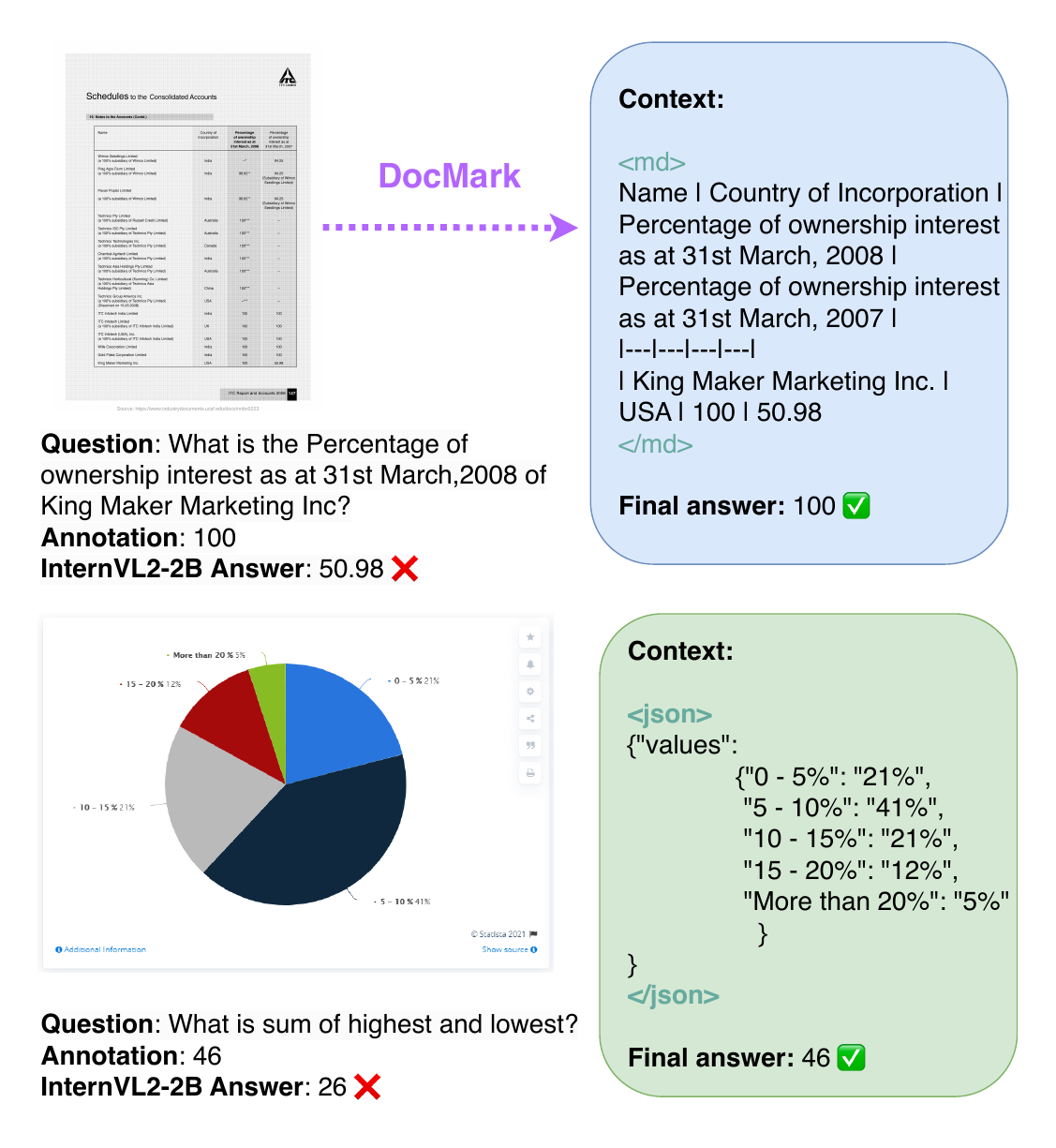}
    \vspace{-0.5em}
\caption{\textbf{Comparison of existing Data annotations and model Predictions: InternVL2-2B vs. our DocMark-2B.} Existing datasets consist primarily of questions and brief answers, often resulting in model hallucinations. In contrast, our model effectively converts documents into structured markup languages, offering critical contextual information that enhances the accurate interpretation of the content.}
    \label{fig:intro}
    \vspace{-2em}
\end{figure}

The core challenge in performing visual document understanding tasks lies in the efficient representation and effective perception of the visual content and its layout information present in the images. However, unlike Large Language Models (LLMs), whose training data generally include comprehensive descriptions and detailed reasoning, common fine-tuning datasets in the MLLM domain usually contain only questions and brief answers.
As shown in \cref{fig:intro}, answers derived from PDF documents often consist of only a short phrase or a single number, even though understanding various parts of the document is required to answer the question. 
Similarly, datasets related to charts often lack a structural interpretation of the image content, resulting in challenges for effective analysis.
Fine-tuning on such data frequently leads to hallucinations and insufficient understanding of the contextual and spatial relationships among various visual elements in the documents. 
Moreover, visual documents come in a wide range of types, exhibiting different semantic content, multiple levels of granularity, and complex layouts. This variety requires adaptive encoding and perception of document images to fully capture their inherent complexity.
These observations raise a critical question: \emph{How can we build contextually grounded training data and devise effective training pipelines to generalize across a wide range of document images?}

In this paper, we propose the adaptive generation of markup languages by the MLLM to facilitate a comprehensive understanding and provision of explicit context for various visual documents.
Document images are often linked to their source codes, termed \textit{markup languages}, which are both well-structured and easy to read. 
Examples include Markdown or LaTeX for PDF documents, HTML for webpages, and TikZ for scientific diagrams—all preserving rich semantic and layout information crucial for describing diverse document types.
Furthermore, their structured nature and clarity make them readily interpretable by pretrained language models, providing valuable contextual information for enhanced document comprehension.
To achieve this goal, we introduce DocMark-Pile, a comprehensive pretraining dataset of 3.8M paired data, designed to parse documents into their corresponding markup languages. This dataset cover a diverse range of visual document types, including natural scene images, dense documents, receipts, LaTeX formulas, web pages, tables, charts, and mathematical diagrams. 
The markup languages extend beyond basic text recognition and grounding, enabling code generation in formats such as Markdown, LaTeX, HTML, JSON, and TikZ.
Building on the pretrained MLLM, we develop an adaptive generation pipeline by finetuning on DocMark-Instruct, a chain-of-thought-like reasoning dataset consisting of 624k annotations. 
Rather than manually specify which markup language to generate by the pretrained model, we let the model itself identify the most suitable type for each document, extracting useful context as intermediate rationales for answering questions.
This approach mimics human-like reasoning, as a person first scans a visual document, then localizes and retrieves contextually relevant information to address specific queries. 
Similarly, when presented with a question, our model generates the most suitable markup languages, such as HTML for webpages, to extract relevant information and formulate answers.
This fine-tuning pipeline enables our model to automatically adapt the appropriate markup language to the document type. Furthermore, it simulates human-like reasoning by disentangling perception from reasoning, encouraging the model to provide essential rationales for accurate answers.
Extensive experiments demonstrate that our models substantially outperform existing state-of-the-art MLLMs on several challenging visual document understanding tasks.

Our contributions can be summarized as follows:

1. We propose a novel pipeline that adaptively uses various markup languages to bridge the gap between visual inputs and linguistic understanding and significantly enhance document comprehension capabilities.

2. We introduce two fine-grained structured datasets, DocMark-Pile and DocMark-Instruct, tailored for pretraining in specialized document parsing and fine-tuning on contextually-grounded instruction following. These datasets enable our model to effectively process complex document formats, including Plain Text, Markdown, LaTeX, HTML, JSON, and TikZ.

3. Extensive experiments demonstrate that our model consistently outperforms state-of-the-art multimodal LLMs across various model sizes, excelling in both markup parsing and downstream reasoning tasks.

\section{Related Work}

\subsection{Multi-modal large language models (MLLMs).}
The field of Natural Language Processing (NLP) has seen remarkable advancements, particularly with the rise of Large Language Models (LLMs)~\cite{OpenAI2023ChatGPT,radford2019language,OpenAI2023GPT4TR}. 
The GPT series by OpenAI~\cite{OpenAI2023ChatGPT} has harnessed the power of massive model scaling, featuring architectures with billions and even trillions of parameters. To enhance instruction-following abilities, models like InstructGPT~\cite{Ouyang2022TrainingLM} and ChatGPT~\cite{OpenAI2023ChatGPT} have been developed, exhibiting exceptional fluency and versatility across various open-domain conversation tasks. Moreover, the instruction tuning based on open-source LLMs~\cite{Touvron2023Llama2O} has gained great popularity.
Motivated by the advancements achieved by LLMs, recent research~\cite{han2023imagebind, lin2023sphinx, shao2024visual} has explored the application of LLMs for multi-modal tasks. Prevalent methods involve adapting the Linear Projector~\cite{llava} or QFormer~\cite{li2023blip} to align vision encoder outputs with the textual feature space of LLMs. Nevertheless, existing methods struggle with comprehensive scene understanding due to limitations in visual representations. To mitigate this, several strategies have been proposed: utilizing composition of multiple vision encoders~\cite{lin2023sphinx, li2024mini}, employing visual encoders designed for high-resolution inputs~\cite{hong2024cogagent}, and implementing dynamic resolution strategies to partition high-resolution images into multiple low-resolution segments~\cite{liu2024llava, li2024monkey,chen2024far}. 
While these strategies have shown advancements, they still struggle with complex understanding and reasoning tasks. This shortcoming arises from the fact that the pretraining and fine-tuning datasets used for MLLM training typically contain limited contextual information and lack a thorough reasoning process.
In this paper, we address this gap by designing an adaptive markup generation pipeline aimed at producing explicit contextual information, which is crucial for effectively performing comprehensive reasoning tasks.

\subsection{Visual Document Understanding}

Recent advancements in multimodal large language models (MLLMs) have significantly enhanced visual document understanding tasks by employing autoregressive training on document images. UReader~\cite{ye2023ureader} introduces a Shape-adaptive Cropping Module to divide high-resolution images into low-resolution sub-images, effectively capturing rich text information. Similarly, Monkey utilizes a sliding window approach to minimize redundancy across different views of the same document. Models such as DocOwl 1.5~\cite{hu2024mplug}, LLaVA-Next~\cite{liu2024llava}, and InternVL~\cite{chen2024far} have explored the utility of OCR pretraining and dynamic partition techniques to address various aspect ratios. LLaVA-Read~\cite{zhang2024llava} integrates bi-resolution images and additional OCR encoders to improve textual information encoding. Other approaches, including DocPedia~\cite{feng2023docpedia} and CogAgent~\cite{hong2024cogagent}, employ diverse encoding strategies to enhance visual representations.
Despite these progress, challenges persist in processing complex document formats, particularly in interpreting their layout information. Additionally, existing document instruction datasets are often limited to questions and short answers, which can lead to hallucinations and restrict deeper contextual understanding. Since visual documents are typically high-structured data tied to their source codes, we develop contextually-grounded training data through markup language parsing and reasoning tasks. By training on structured document representations, our method aims to leverage the code interpretation capabilities of LLMs to further enhance visual document understanding.
\section{Method}

\subsection{DocMark-Pile: Multi-Task Pretraining for Markup Language Conversion}
\begin{figure*}[t]
    \vspace{-2em}
    \centering
    \includegraphics[width=0.97\linewidth]{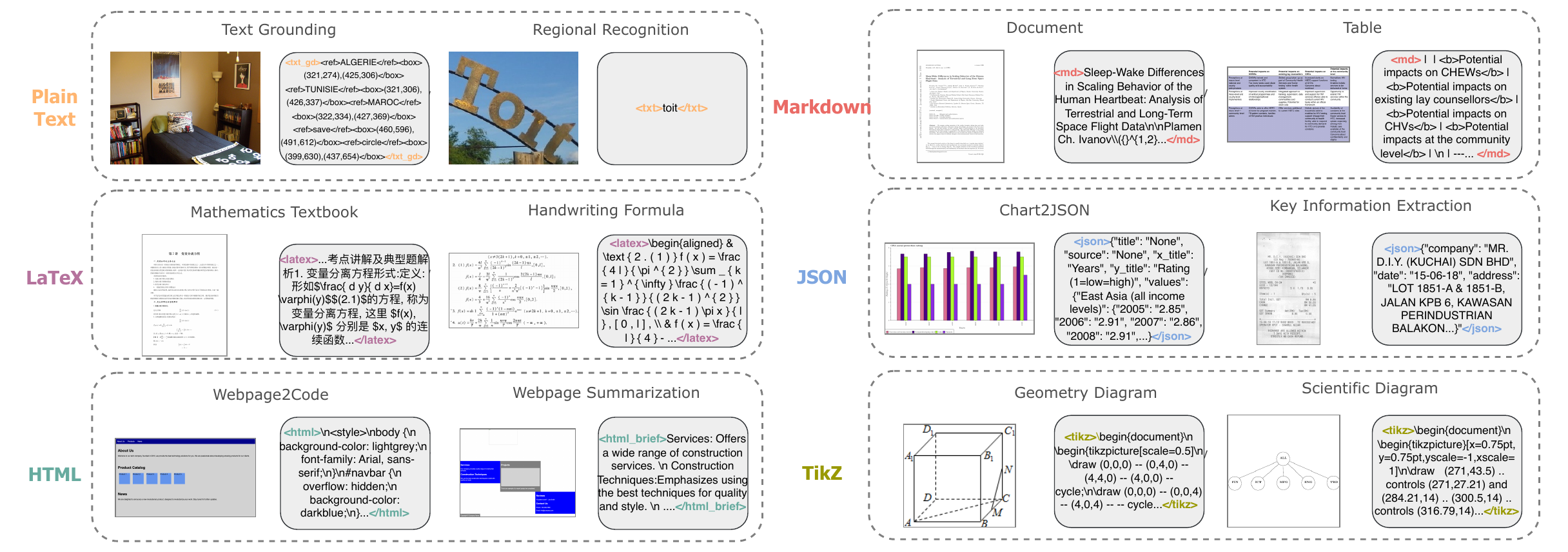}
    \vspace{-0.5em}
\caption{\textbf{Overview of our DocMark-Pile dataset for pretraining on various markup language generation tasks.} Due to the length of the answer, we have omitted a part of the content for better display.}
    \label{fig:datavis}
    \vspace{-1em}
\end{figure*}

\begin{figure}[t]
    \centering
    \includegraphics[width=0.95\linewidth]{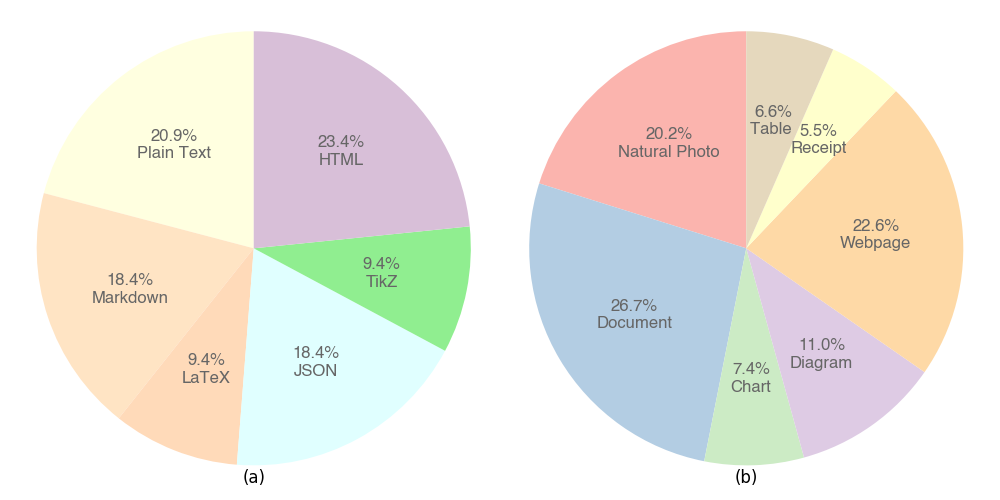}
    \vspace{-0.5em}
\caption{\textbf{Domain distribution of our DocMark-Pile dataset.} (a) Distributions of different markup language parsing tasks. (b) Distributions of different image types.}
    \label{fig:distribution}
    \vspace{-2em}
\end{figure}
Visual document understanding encompasses various tasks, such as comprehending paragraphs from PDF documents, interpreting numerical data in charts and tables, and retrieving information from receipts, among others. 
Most documents are generated from source code or can be converted into structured representations. 
These structured formats act as markup languages, which define the structure of various elements, helping to convey meaning and organize the content effectively.
To enhance the model's ability to understand and interpret a wide range of visual document types, we propose a multi-task pretraining dataset, DocMark-Pile, which focuses on translating documents into their corresponding markup languages. 
To ensure comprehensive coverage of document understanding scenarios, we utilize six distinct markup languages: plain text, Markdown, LaTeX, HTML, JSON, and TiKZ, as demonstrated in \cref{fig:datavis}. 
Each format serves a unique purpose, which we will elaborate on below.
For data acquisition, we utilize publicly available datasets as well as synthetic methods while maintaining the consistency between images and markup languages.

\subsubsection{Specialized Task Design}
\textbf{Document to TXT } Plain Text is the simplest format for basic content extraction from documents, as it retains all the textual information contained in the image. We primarily adopt this format for natural photos and regional text images, as these types of images typically contain straightforward textual information that can be extracted without formatting complexities. 
For natural photos, we utilize public datasets including TextOCR~\cite{singh2021textocr}, COCO-Text~\cite{veit2016coco}, HierText~\cite{long2022towards}, LSVT~\cite{sun2019icdar}, RCTW-17~\cite{shi2017icdar2017}, and CTW1500~\cite{yuliang2017detecting}. These datasets cover a wide range of scenarios, including advertising posters, signage, product labels, and image subtitles. Additionally, they often include bounding box annotations for the target text. We leverage these annotations to formulate text grounding tasks that require the model to recognize both the text and its coordinates within the image.
For regional text images, we further incorporate datasets such as IC13~\cite{karatzas2013icdar}, IC15~\cite{karatzas2015icdar}, Union14ML~\cite{jiang2023revisiting}, and WordArt~\cite{xie2022toward}. These images consist exclusively of regions containing text, which are usually in the form of single lines or curved text.
We standardize the training data format by reading the text in the order from left to right and top to bottom, including additional bounding box coordinates for the text grounding task.

\textbf{Document to Markdown}
Markdown is a lightweight syntax which enables basic formatting options such as bold, italics, headings, and lists. Markdown can be easily converted to and from other formats like HTML and LaTeX.
We employ the Markdown format for dense text documents and tables because it effectively represents their structured content.
For dense text documents, we primarily source our data from arXiv, where we obtain articles along with their corresponding source code and compiled PDFs.
Following the process established in~\cite{hong2024cogagent}, we convert the PDFs into images while simultaneously translating the LaTeX code into Markdown format.
Since many articles include complex formulas that cannot be successfully compiled in Markdown, we filter these instances during the conversion process.
For tables, we draw from publicly available datasets, including PubTabNet~\cite{zhong2020image}, SynthTabNet~\cite{nassar2022tableformer}, TabRecSet~\cite{yang2023large}, TableGraph~\cite{xue2021tgrnet}, and SciTSR~\cite{chi2019complicated}. We select these datasets because their annotations contain the original table content. Since many of these annotations are provided in HTML format, we convert them into Markdown format to meet the requirements of our task.

\textbf{Document to LaTeX}
LaTeX allows for the precise modeling of academic papers, particularly due to its capability to typeset complex mathematical formulas and scientific notation. We gather images converted from mathematics and science textbooks and exam papers, since these images typically contain numerous formulas and equations that must be accurately represented in LaTeX format. 
Subsequently, we utilize Mathpix to generate annotations for these documents.
In addition, we incorporate handwritten mathematical expression recognition datasets, including HME100K~\cite{yuan2022syntax} and MathWriting~\cite{gervais2024mathwriting}, which convert handwritten mathematical formulas into LaTeX format. 
We further enrich these data sources with additional rendered images. We first extract original mathematical expressions from our collected corpora, then use the Python library Matplotlib to render them as formula images. Subsequently, we apply data augmentation techniques, such as affine transformations and the addition of Gaussian noise, to enhance the robustness of the model in recognizing images under various backgrounds and display conditions.

\textbf{Document to HTML}
For webpage understanding, we introduce the task of parsing webpages into HTML formats. We utilize WebSight~\cite{laurenccon2024unlocking}, a synthetic dataset consisting of HTML codes and their corresponding screenshots. Additionally, we augment our dataset by scraping real website images and HTML codes.
Since many real website screenshots contain sparse content, directly learning from their HTML code often fails to extract useful information due to the noise. 
To address this, we introduce a webpage summarization task, which focuses on identifying and interpreting the key points and core content within a webpage. 
Specifically, we extract only the main elements of the page, such as the title, content, and other essential information.
By focusing on the summary of key elements, this approach reduces the complexity and noise inherent in real website data, enhancing the quality of the pretraining data.

\textbf{Document to JSON} To facilitate structured information extraction, we utilize JSON (JavaScript Object Notation) format, which represents information as key-value pairs, enhancing both readability and conciseness. 
This extraction is particularly valuable in scenarios involving cards, receipts, and forms, where summarizing and retrieving important data such as names, dates, and addresses from images is critical.
To achieve this, we leverage several public datasets, including CORD~\cite{park2019cord}, POIE~\cite{kuang2023visual}, SVRD~\cite{yu2023icdar}, WildReceipt~\cite{sun2021spatial}, and XFUND~\cite{xu2022xfund}, whose annotations can be easily converted into JSON format. 
In addition to textual information, we incorporate structured data from charts.
We represent this information in a Python dictionary that encompasses fields such as "title", "source", "x-axis", "y-axis", and "value". For chart-related data, we utilize the PlotQA~\cite{methani2020plotqa} dataset, which contains meta-information about chart content.
Moreover, we augment our dataset by rendering various types of chart images. This involves randomly creating textual labels and numerical values. Then we employ Matplotlib to produce diverse chart images, including bar plots, scatter plots, line plots, and dot plots, using randomly sampled fonts and colors.

\textbf{Document to TikZ}
TikZ is a markup language commonly used for creating scientific figures. To construct our dataset, we leverage the open-source training sets of DaTikZv1~\cite{belouadi2024automatikz} and DaTikZv2~\cite{belouadi2024detikzify}, which contain TikZ graphics collected from arXiv papers and artificial examples, along with their corresponding TikZ programs.
Additionally, we extract scientific drawings from a curated corpus of mathematical and scientific textbooks and employ a trained model from~\cite{belouadi2024detikzify} to convert these illustrations into TikZ code. 
Recognizing that conversion may not always be accurate, we take the necessary step of recompiling the TikZ code and use the re-generated images in our training. We also exclude any instances that fail to compile, ensuring that the TikZ code remains consistent with the figures.

\subsubsection{Dataset Statistics}

DocMark-Pile, a comprehensive dataset for document understanding, comprises 3.8 million examples of diverse markup language parsing tasks and image types.  

\textbf{Markup Language Categories:} We visualize the distribution of different language parsing tasks in~\cref{fig:datavis}, showcasing the percentage of each task type. This highlights the diversity of our dataset.
In our data annotation, we prepend the special tokens such as \texttt{<md>} and \texttt{</md>} or \texttt{<json>} and \texttt{</json>} to the beginning and end of answers to indicate the corresponding markup language category.

\textbf{Image Types:}  Our dataset covers a wide range of image types, encompassing a comprehensive representation of visual documents. For clarity, we categorize text-dense materials like academic papers, textbooks, exam papers, and receipts under "documents", which constitutes the largest portion of the dataset.

\begin{figure*}[t]
    \vspace{-2em}
    \centering
    \includegraphics[width=0.95\linewidth]{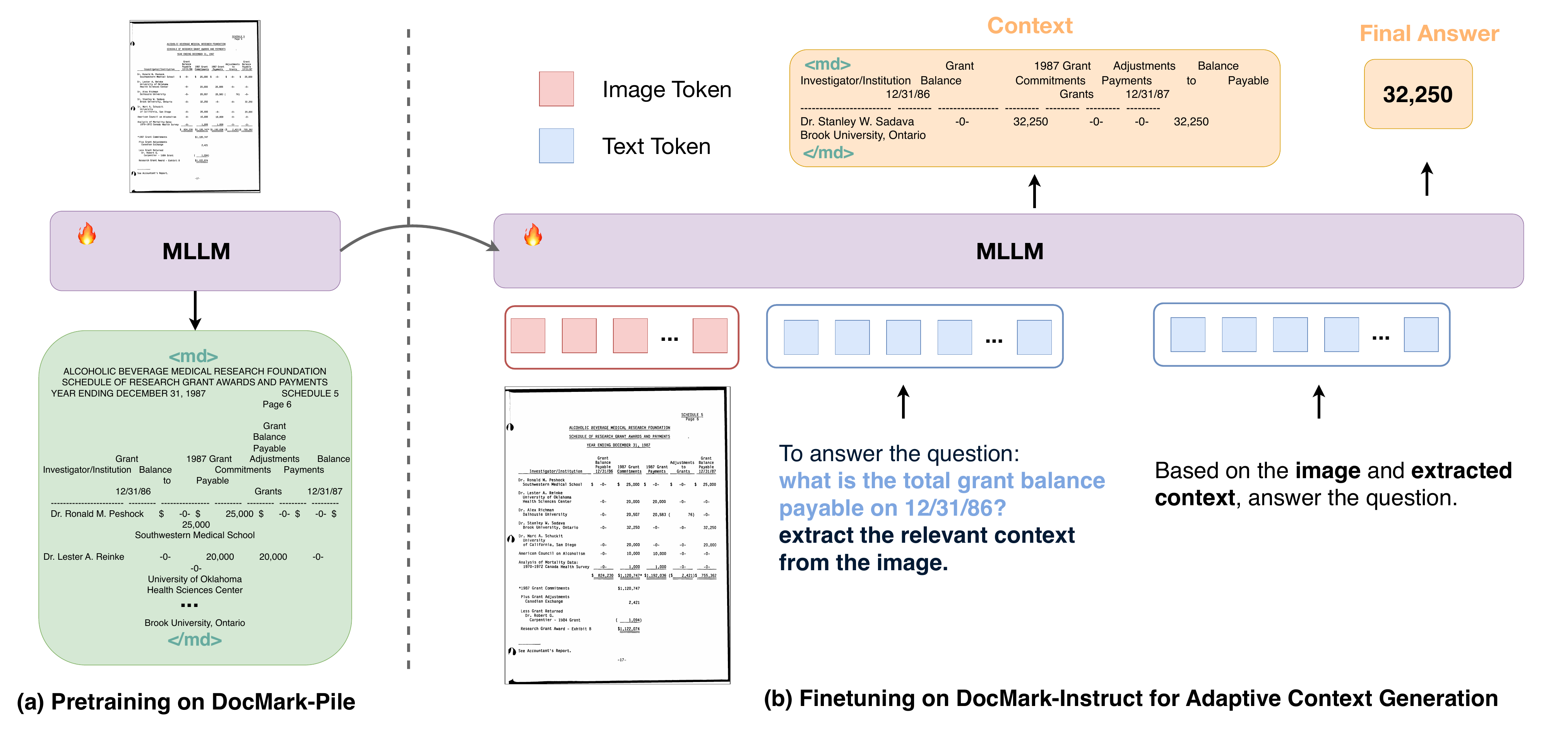}
    \vspace{-0.5em}
\caption{\textbf{Overview of our adaptive markup language generation framework.} (a) During pretraining, the model learns to parse documents into corresponding markup language representations, enhancing its structural understanding capabilities. (b) In the finetuning phase, the model is optimized to identify and extract relevant information as intermediate rationales to formulate precise answers.}
    \label{fig:pipeline}
    \vspace{-1.2em}
\end{figure*}
\subsection{Adaptive Markup Language Generation During Fine-tuning}

\subsubsection{Adaptive Generation Pipeline}
Through multi-task pretraining for versatile markup language generation, our model gains enhanced abilities to understand both visual content and layout information in diverse documents.
We further fine-tune the model using instruction-tuning datasets, equipping it with the skills to handle downstream tasks such as open-ended question-answering related to document contexts.
Previous methods primarily utilize questions and brief answers from public benchmark training sets, limiting their ability for deeper comprehension.
In contrast, our model's ability to generate structured markup languages offers a significant advantage. These generated markup languages can serve as auxiliary context cues that facilitate more complex reasoning tasks. 
Being highly concise and structured, markup languages encode textual information efficiently and can be processed effectively by language models pre-trained on extensive text and code corpora.

Motivated by chain-of-thought reasoning widely employed in the domain of LLMs, we design an adaptive markup language generation pipeline to enhance reasoning abilities. 
Rather than manually determining the markup language type to generate by the pretrained model as a prefix, we let our model automatically generate the appropriate markup language as an intermediate step to derive answers.
This mirrors the human-like chain-of-thought reasoning; when reading documents such as PDFs or textbook pages, humans first scan the content, then localize and retrieve relevant information to answer specific questions.

In our adaptive pipeline, the model automatically determines the appropriate markup language type and identifies contextually relevant information to answer the question. 
We implement this as a two-round conversational process as follows:
\begin{align}
\label{cot}
\text{ $Q_1$: } & \text{To answer the question: $<Question>$,} \nonumber\\
&\text{extract the relevant context from the image.} \nonumber\\
\text{ $A_1$: } & \text{$<type><Context></type>$} \nonumber \\
\text{ $Q_2$: } & \text{Based on the image and extracted context,} \\
&\text{answer the question: $<Question>$} \nonumber\\
\text{ $A_2$: } & \text{$<Answer>$} \nonumber
\end{align}
In this pipeline, the model is first asked to provide the essential contextual information to address the specific question $<Question>$. It then outputs this context as $<Context>$, serving as an intermediate rationale in the process.
 Special tokens $<type>$ and $</type>$ indicate the markup language type.
 In the second round, the model derives the final answer by utilizing both the image and the context provided in the first round.
This structured approach ensures that the model systematically processes and retains critical contextual information, thereby enhancing its ability to generate accurate and contextually grounded answer akin to human thought processes.

\subsubsection{DocMark-Instruct: CoT Reasoning Data Construction}
\label{sec:cot}
To enable our model with adaptive generation capabilities, it is essential to curate a dataset formatted as described in template (\ref{cot}).
However, existing fine-tuning datasets only contain question-answer pairs and lack the detailed CoT reasoning process. To address this gap, we propose an efficient method for constructing the required data, \textbf{DocMark-Instruct}.
Most questions in common training datasets within the document domain are closely related to the textual information present in the images. We observe that these questions can often be answered directly by providing the corresponding markup language to a powerful language model, thus eliminating the need for visual input.
Therefore, we first utilize our pretrained model to convert images into their corresponding markup languages. Next, we employ ChatGPT-3.5, a freely accessible large language model known for its robust language understanding and generation capabilities, to create additional annotations. We input the question-answer pairs and the entire markup language text into it, asking it to retrieve the necessary information to answer the questions, finally producing 624k annotations.
Our data construction pipeline circumvents the high annotation costs typically associated with proprietary MLLMs such as GPT4-V, making it a practical and accurate solution for developing comprehensive CoT reasoning datasets.

\subsubsection{Training and Inference}

During the fine-tuning phase, we concatenate the two-round dialogue data for the forward process and apply Cross-Entropy loss on both answers, $A_1$ and $A_2$. 
In the inference phase, we first input $Q_1$ to prompt the model to generate the adaptive markup language context. Following this, we formulate $Q_2$ to obtain the final response.
Importantly, our training pipeline is also compatible with common training data that lacks the CoT reasoning process. Users can directly ask the model to answer a question by omitting the CoT prompt described in template (\ref{cot}).

\section{Experiments}
\subsection{Implementation Details}

\setlength{\tabcolsep}{5pt}
\begin{table*}[]
\adjustbox{max width=0.95\textwidth}{%
\begin{tabular}{lcccccccc}
\toprule
\multirow{2}{*}{Method} &  \multirow{2}{*}{Text Recognition} & \multirow{2}{*}{HME100K} & \multicolumn{3}{c}{ChartQA-SE} &   \multicolumn{2}{c}{KIE}& \multirow{2}{*}{DaTikZ-test} \\ 
\cmidrule(lr){4-6} \cmidrule(lr){7-8} 
 &  &  & AP@strict & AP@slight &   AP@high & FUNSD & SROIE \\ 
\cmidrule(lr){1-1}\cmidrule(lr){2-9}

Paddle OCR & \cellcolor{gray!30} 140  &  \\
{CAN-ABM~\cite{li2022counting}} &  & \cellcolor{gray!30} 64.3  &  \\
{OneChart~\cite{chen2024onechart}} &  & & \cellcolor{gray!30}72.0  &  \cellcolor{gray!30} 82.9 &\cellcolor{gray!30} 85.9 \\
{Nougat~\cite{blecher2023nougat}} &  & &   & & & \cellcolor{gray!30} 55.4 & \cellcolor{gray!30} 33.6 \\
{DT-DS$_{1.3B}$~\cite{belouadi2024detikzify}} &  & &   & & &  &  & \cellcolor{gray!30} 0.22  \\
{DT-CL$_{7B}$~\cite{belouadi2024detikzify}} &  & &   & & &  &  & \cellcolor{gray!30} \textbf{0.68}  \\

\cmidrule(lr){1-1}\cmidrule(lr){2-9}

{DocMark-2B}  & 247 & 67.2 &  73.8 & 85.2 & 86.5 & 64.1 & 54.2 & 0.39  \\
{DocMark-8B}  & \textbf{252} & \textbf{73.4} &  \textbf{75.4} & \textbf{85.5} & \textbf{87.1} & \textbf{68.5} & \textbf{59.6} & {0.64}  \\
\bottomrule
\end{tabular}
} 
\caption{\textbf{Performance comparison with expert models on specialized tasks.} The best results are marked in bold.}
\vspace{-1.3em}
\label{tab:ref}
\end{table*}

\textbf{Model Architecture} We adopt the InternVL~\cite{chen2024internvl} architecture for our model due to its enhanced visual representation capabilities. We use the pretrained InternViT~\cite{chen2024internvl} as the visual encoder and utilize InternLM-2B~\cite{team2023internlm} and InternLM-8B~\cite{team2023internlm}, known for their robust linguistic understanding abilities. We adopt MLP layers to project the image into visual tokens, which are further concatenated with text tokens. As our work focuses on incorporating adaptive markup language generation to improve visual document understanding, we do not explore additional resolution scaling strategies in this paper. Instead, we employ the dynamic resolution method proposed in InternVL2~\cite{chen2024far}, which demonstrates efficacy in modeling high-resolution document images. Specifically, the input image is split into several sub-images of resolution \(448 \times 448\), with a maximum of 12 sub-images used in our training settings.

\textbf{Training Settings} Our model is trained through a two-stage process. In the first stage, we utilize DocMark-Pile to empower the model with markup language parsing abilities. Additionally, we incorporate the image captioning dataset LLaVA-558K~\cite{llava} to ensure alignment between the visual input and text tokens. We pre-train the model for one epoch with a learning rate of 2e-5 and a total batch size of 128.
In the second stage, we fine-tune our model on the proposed CoT Reasoning dataset DocMark-Instruct. As described in~\cref{sec:cot}, we create additional CoT annotations for the training sets of TextVQA~\cite{textvqa}, DocVQA~\cite{mathew2021docvqa}, InfographicsVQA~\cite{mathew2021infographicvqa}, ChartQA~\cite{masry-etal-2022-chartqa}, DVQA~\cite{kafle2018dvqa}, XFUND~\cite{xu2022xfund}, and Geo170K~\cite{gao2023gllavasolvinggeometricproblem}. Apart from DocMark-Instruct, we also integrate general instruction tuning data from ShareGPT4V~\cite{chen2023sharegpt4v} and text-only multi-turn conversation datasets~\cite{zhou2024lima,kopf2024openassistant}. This integration preserves the model's capacity for general understanding and conversational interaction. We fine-tune our model on these datasets for one epoch with a learning rate of 1e-5 and a total batch size of 128.

\subsection{Evaluation on Markup Language Generation Abilities}

\textbf{Evaluation Settings} The ability to generate accurate markup languages is crucial for our model to understand complex document formats and provide adaptive context for facilitating downstream task reasoning. 
We comprehensively evaluate our model on specialized tasks and compare it with domain-expert models. 
We utilize the text recognition benchmark in~\cite{liu2023hidden}, which covers a broad range of scene texts.
We also assess our model's accuracy in expressing formulas present in images using the LaTeX format, utilizing the test set of HME100K~\cite{yuan2022syntax}.
Moreover, we adopt the ChartQA-SE benchmark proposed in~\cite{chen2024onechart} and report AP@strict, AP@slight, and AP@high metrics. We also evaluate on FUNSD and SROIE for key information extraction task. These metrics measure the accuracy of the structural extraction results compared to the ground-truth values.
For TiKZ code generation, we use the DaTikZ test set proposed in~\cite{belouadi2024detikzify}. We report the average of image similarity and code similarity to assess the code consistency.
By thoroughly evaluating our model across these diverse tasks, we aim to demonstrate its capability to accurately generate markup languages, thereby facilitating effective visual document understanding and reasoning.

 \textbf{Results.}
We present the comparison results of our model against various expert models in~\cref{tab:ref}.
As can be seen, our DocMark models outperform the expert models by a significant margin on the text recognition and HME100K~\cite{yuan2022syntax} benchmarks. 
Moreover, our model demonstrates impressive capabilities in structural information extraction, particularly on the ChartQA-SE~\cite{chen2024onechart} and KIE tasks.
On DaTiKZ-test~\cite{belouadi2024detikzify}, DocMark-8B achieves performance nearly on par with DT-CL$_{7B}$, without additional search costs for TiKZ code generation.
These results showcase the model's effectiveness in markup language conversion, enhancing visual perception and supporting contextually grounded reasoning.
\begin{table*}[t]
\adjustbox{max width=0.98 \textwidth}{%
\begin{tabular}{lcccccccccc}
\toprule
\multicolumn{1}{c|}{Method} & \multicolumn{1}{c|}{LLM}  &TextVQA$_{val}$ & DocVQA$_{val}$ & InfoVQA$_{val}$ & ChartQA & AI2D & OCRBench & WebQA  & MathVision\\ 
\cmidrule(lr){1-1}\cmidrule(lr){2-2}\cmidrule(lr){3-10}{\textit{Lightweight Specialist Models}}\\

\cmidrule(lr){1-1}\cmidrule(lr){2-2}\cmidrule(lr){3-10}

\multicolumn{1}{l|}{Donut} & -  & 43.5  & 67.5 & 11.6 & 41.8 & 30.8 & - & - \\
\multicolumn{1}{l|}{Pix2Struct-Base} & -  &-  & 72.1 & 38.2 & 56.0 & 40.9 & - & - & -\\
\multicolumn{1}{l|}{Pix2Struct-Large} & -  & - & 76.6 & 40.0 & 58.6 & 42.1 & - & - & -\\

\cmidrule(lr){1-1}\cmidrule(lr){2-2}\cmidrule(lr){3-10}{\textit{Multimodal LMMs}}\\

\cmidrule(lr){1-1}\cmidrule(lr){2-2}\cmidrule(lr){3-10}
\multicolumn{1}{l|}{Vary-toy} & 1.8B & -  & 65.6 & - & 59.1 & - & - & - & - \\
\multicolumn{1}{l|}{MiniCPM-V-2} & 2.8B & 74.1  & 69.7 & 37.9 & 59.8 & 62.6 & 594 & - & 15.1  \\
\multicolumn{1}{l|}{InternVL2} & 2B  & 73.2  & 86.0 & 57.6 & 76.4  & 74.2 &  783 & 66.5 & 15.7 \\

\cmidrule(lr){1-1}\cmidrule(lr){2-2}\cmidrule(lr){3-10}
\multicolumn{1}{l|}{DocMark (Ours)} & 2B  &   \textbf{74.8} & \textbf{87.8} & \textbf{61.2} & \textbf{79.8} &  \textbf{82.5} & \textbf{813} &  \textbf{70.1}  & \textbf{18.8}\\
\cmidrule(lr){1-1}\cmidrule(lr){2-2}\cmidrule(lr){3-10}
\multicolumn{1}{l|}{Ureader} & 7B  & 57.6  & 80.7 & 46.4 & 70.0 & - & - & - & -\\
\multicolumn{1}{l|}{DocPedia} & 7B  & 60.2  & 47.1 & 15.2 & 46.9 & - & - & - & -\\
\multicolumn{1}{l|}{Vary} & 7B  & -  & 76.3 & - & 66.1 & - & - & - & -\\
\multicolumn{1}{l|}{Monkey} & 9B  & 67.6  & 66.5 & 36.1 & 65.1 & 62.6 & 514 & -  & - \\
\multicolumn{1}{l|}{TextMonkey} & 9B  & 64.3  & 66.7 & 28.6 & 59.9 & - & 558 & - & -\\
\multicolumn{1}{l|}{CogAgent} & 17B  & 76.1  & 81.6 & 44.5 & 68.4 & - & - & - & - \\
\multicolumn{1}{l|}{DocOwl2} & 8B  & 66.7  & 80.7 & 46.4 & 70.0 & - & - & - & -\\
\multicolumn{1}{l|}{InternVL2} & 8B  &  77.4 & \textbf{90.8} & \textbf{72.8} & 82.2 & 83.7 & 794 & 77.5 & 20.2 \\

\cmidrule(lr){1-1}\cmidrule(lr){2-2}\cmidrule(lr){3-10}

\multicolumn{1}{l|}{DocMark (Ours)} & 8B & \textbf{78.0} & 89.8 & 68.3 & \textbf{84.2} & \textbf{86.2} & \textbf{823} & \textbf{78.9} & \textbf{21.1} \\

\bottomrule
\end{tabular}
} 
\vspace{-0.5em}
\caption{\textbf{Performance comparisons on downstream document understanding tasks.} The best results are marked in bold.}
\vspace{-1.5em}
\label{table:downstream}
\end{table*}

\subsection{Evaluation on Downstream Understanding tasks}
\textbf{Evaluation Settings}
To validate the effectiveness of adaptive markup language generation, we evaluate the model's performance on a range of downstream tasks for document understanding. We adopt diverse evaluation benchmarks that encompass a wide variety of multi-modal scenarios.
We report accuracy on TextVQA~\cite{textvqa} to assess the model's ability to answer questions related to natural scene text. 
Additionally, we evaluate performance on DocVQA~\cite{mathew2021docvqa}, InfoVQA~\cite{mathew2021infographicvqa}, AI2D~\cite{kembhavi2016diagram}, and ChartQA~\cite{masry-etal-2022-chartqa} for comprehension of text-dense documents and charts.
To assess the model's general OCR abilities, we utilize OCRBench~\cite{liu2023hidden}, a comprehensive benchmark containing various text-related images. 
For webpage understanding, we report the WebQA accuracy as proposed in VisualWebBench~\cite{liu2024visualwebbenchfarmultimodalllms}.
Furthermore, we evaluate the model's math reasoning performance using MathVision~\cite{wang2024measuring} dataset to validate its ability to interpret complex mathematical diagrams.

\textbf{Results} 
We present the comparison results on downstream understanding tasks in ~\cref{table:downstream}. Our DocMark models consistently outperform state-of-the-art multimodal large language models across nearly all benchmarks, both for the 2B and 7B parameter sizes. Notably, DocMark significantly enhances performance on common document understanding tasks, while also achieving remarkable results on MathVision, demonstrating its effectiveness in addressing complex multimodal math problems. Furthermore, our models show significant improvements in general OCR and webpage understanding, as highlighted by the results on OCRBench and WebQA. When compared to state-of-the-art models like the InternVL2 series, DocMark achieves superior accuracy with substantially less training data. These results illustrate that the proposed framework greatly enhances document understanding and reasoning capabilities.

\begin{figure}[t]
    \centering
     \vspace{1.2em}
    \includegraphics[width=0.9\linewidth]{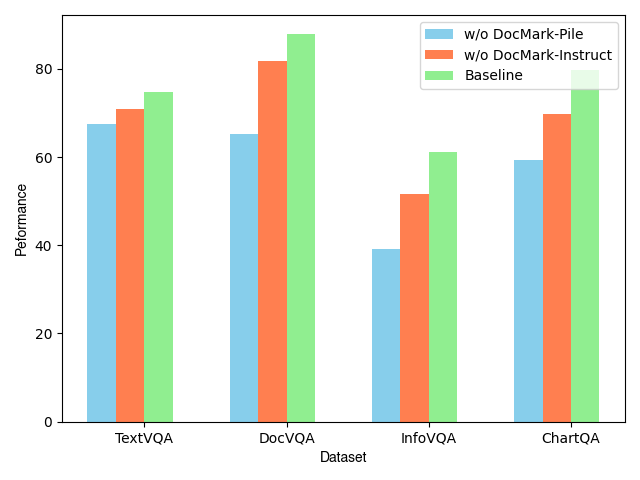}
    \vspace{-1em}
\caption{\textbf{Ablation study on employing different training dataset strategies.}}
    \label{fig:ablation}
    \vspace{-1.2em}
\end{figure}
\subsection{Ablation Study}
To investigate the impact of proposed datasets, we examine various training dataset strategies using DocMark-2B. We train the model without incorporating DocMark-Pile and DocMark-Instruct, relying solely on image-caption datasets and general question-answering datasets for fine-tuning. We report the accuracy of the model on four document-related QA datasets for analysis. Subsequently, we integrate DocMark-Pile for model pretraining and employ DocMark-Instruct for fine-tuning. As illustrated in~\cref{fig:ablation}, the absence of DocMark-Pile significantly degrades model performance, particularly for text-dense documents and structured data. Furthermore, incorporating DocMark-Instruct substantially enhances performance when built upon DocMark-Pile pretraining, demonstrating its effectiveness in improving reasoning capabilities.
We provide more performance analysis in our supplementary.

\section{Conclusion}
In this paper, we have introduced an innovative framework that utilizes adaptive markup language generation to enhance visual document understanding. Our approach addresses the core challenge of efficiently representing and perceiving visual content and layout information in images by incorporating specialized pretraining tasks and an adaptive generation pipeline. The proposed method not only improves the versatility of MLLMs in handling complex document formats, but also enables them to perform chain-of-thought reasoning to produce contextually grounded answers. Extensive experiments demonstrated that our models outperform existing state-of-the-art approaches on several challenging visual document understanding tasks.

\section*{Acknowledgments}
This project is funded in part by National Key R\&D Program of China Project 2022ZD0161100, by the Centre for Perceptual and Interactive Intelligence (CPII) Ltd under the Innovation and Technology Commission (ITC)'s InnoHK, by General Research Fund of Hong Kong RGC Project 14204021, by NSFC-RGC Joint Fund Project N\_CUHK498/24, by the National Natural Science Foundation of China (No.62206272). Hongsheng Li is a PI of CPII under the InnoHK.
{
    \small
    \bibliographystyle{ieeenat_fullname}
    \bibliography{main}
}

\clearpage
\setcounter{page}{1}
\setcounter{section}{0}
\renewcommand{\thesection}{\Alph{section}}
\maketitlesupplementary
\section{Details of DocMark-Instruct Dataset}
We provide a detailed overview of the DocMark-Instruct dataset, which is designed to enhance the document understanding and context grounding capabilities of MLLMs in document-related scenarios.
The dataset contains data from six distinct domains, including Text-based QA, Document-based QA, Chart-based QA, Key Information Extraction, Webpage-based QA and Mathematical QA.
To enrich the dataset, we make use of multiple public datasets.
Additionally, we also include in-house data created through careful curation.
The domain distribution and statistics of our dataset are shown in \cref{tab:cot}.

\section{Prompt Design}
\subsection{Prompt for Different Pretraining Tasks on DocMark-Pile}

For multi-task pretraining on DocMark-Pile, we utilize a variety of instruction prompts specifically designed for each markup language generation task. These prompts not only guide the model during pretraining but also enhance its versatility in tackling diverse markup translation challenges. Some examples of the prompts utilized during pretraining are listed in ~\cref{tab:prompt}.

\subsection{Prompt for Creating the DocMark-Instruct Dataset}
\begin{figure*}[t]
    \vspace{0em}
    \centering
    \includegraphics[width=0.97\linewidth]{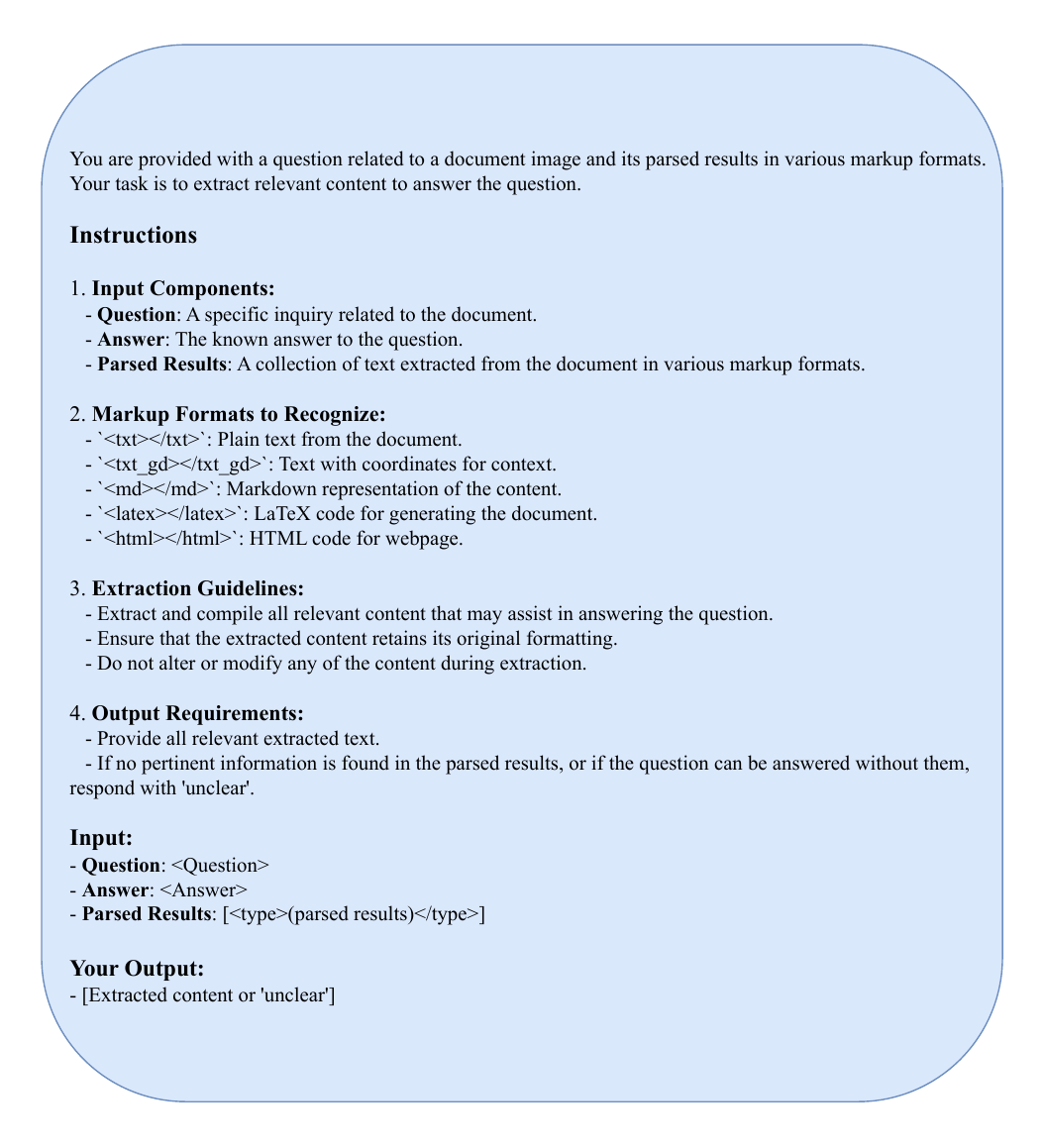}
    
\caption{\textbf{Prompt for creating the DocMark-Instruct dataset.}}
    \label{fig:prompt_chatgpt}
    
\end{figure*}

As outlined in Section 3.2.2 of the main paper, we employ ChatGPT-3.5 to generate immediate context for our DocMark-Instruct dataset. Specifically, we prompt the model to extract relevant information from the provided markup language necessary for answering the questions. 
It is important to note that some questions may not relate directly to the textual information; for instance, certain questions may only require general knowledge for answers. In such cases, we instruct the model to respond with \texttt{"unclear"}, indicating that the textual information is not applicable. This approach helps prevent inaccurate annotations and mitigates the risk of model hallucinations.
The detailed prompt template is provided in~\cref{fig:prompt_chatgpt}.

\begin{table}[t]
\centering
\adjustbox{max width=0.5\textwidth}{%
\begin{tabular}{lccc}
    \toprule
    \textbf{Task} & \textbf{Source Dataset} & \textbf{Markup Type} & \textbf{\#Num.} \\
    \midrule
     \multirow{3}{*}{Text-based QA} & TextVQA~\cite{textvqa} & $txt,txt\_gd$ & 29k \\
    & STVQA~\cite{biten2019scene} &  $txt,txt\_gd$ & 26k \\
    & EST-VQA~\cite{wang2020general} & $txt,txt\_gd$ & 8k  \\
    \midrule
    \multirow{3}{*}{Document-based QA} & DocVQA~\cite{mathew2021docvqa} & $md$ & 31k \\
    & InfoVQA~\cite{mathew2021infographicvqa} & $md$ & 12k \\
    & Docmatix~\cite{Docmatix} & $md$ & 50k \\
    & In-house data & $md$, $latex$ & 12k \\
    \midrule
    \multirow{3}{*}{Chart-based QA} & ChartQA~\cite{masry-etal-2022-chartqa} & $json$ & 44k \\
    & PlotQA~\cite{methani2020plotqa} & $json$ & 146k \\
    & DVQA~\cite{kafle2018dvqa} & $json$ & 77k \\\midrule
      \multirow{4}{*}{Key Information Extraction} & POIE~\cite{kuang2023visual} & $json$ & 2k \\
    & SROIE~\cite{huang2019icdar} & $json$ & 0.6k \\
    & FUNSD~\cite{jaume2019funsd} & $json$ & 0.1k \\
    & XFUND~\cite{xu2022xfund} & $json$ & 0.2k \\\midrule  
\multirow{3}{*}{Webpage-based QA} 
 & WebSRC~\cite{chen2021websrc} & $html$ & 59k \\
 & In-house data & $html$ & 4k \\\midrule
\multirow{3}{*}{Mathematical QA} & Geo170k~\cite{gao2023g} & $tikz$ & 48k \\
& Geometry3k~\cite{lu2021inter} & $tikz$ & 2k \\
 & MultiMath~\cite{peng2024multimath} & $tikz$ & 50k \\
  & In-house data & $tikz$ & 17k \\\midrule
    Total & - & - & 624k \\
    \bottomrule
    \end{tabular}
} 

\label{tab:cot}
\caption{\textbf{Overview of the DocMark-Instruct Dataset.}}
\end{table}
\vspace{-2em}

\newpage

\begin{table*}[t]
\centering

\adjustbox{max width=\textwidth}{%
\begin{tabular}{ll}
    \toprule
    \textbf{Task} & \textbf{Prompt}  \\
    \midrule
    \multirow{3}{*}{Text Recognition} & Kindly recognize the text from the image.\\
     & How can I extract the text from the image? \\
    & What text is in the image that can be extracted? \\ \midrule
    
    \multirow{3}{*}{Text Grounding} &  Can you perform text extraction with grounding? \\
    & Please detect the text with grounding from the image. \\
    & Recognize the text with grounding from the image. \\ \midrule

     \multirow{3}{*}{Image to Markdown} & Parse the image into a proper markup language format. \\
      & How to convert text from the image to markdown format? \\
      & How to extract text from the image and change it to markdown format? \\ \midrule

     \multirow{3}{*}{Image to LaTeX} & Convert the image into a structured format. \\
      & How to extract and translate text from the image to LaTeX format? \\
      & How can I convert text from the image to LaTeX format efficiently? \\ \midrule
    \multirow{3}{*}{Image to HTML} & What is the HTML code corresponding to this image? \\
      & Generate the HTML code. \\
      & Parse the image into an appropriate markup language format. \\ \midrule
    \multirow{3}{*}{Webpage Summarization} &  What is the main idea of this webpage screenshot? \\
      & What are the main information points of the webpage shown in the image? \\
      & What is the key message conveyed by this webpage image? \\ \midrule
    \multirow{3}{*}{Image to JSON} & Extract text from the image in JSON format. \\
      & Output the image text as JSON. \\
      & Represent the image text in a structured format. \\ \midrule
    \multirow{3}{*}{Image to TikZ} & I need to get the code for drawing this image. \\
      & What is the TikZ code for this image?. \\
      & Please show me the TikZ code for displaying this image. \\
    \bottomrule
    \end{tabular}
} 
\caption{\textbf{Examples of prompts for different pretraining tasks on DocMark-Pile.}}
\label{tab:prompt}
\end{table*}

\begin{table*}[t]

  \centering
    \setlength{\tabcolsep}{0.02\linewidth}
  \adjustbox{max width=0.9\textwidth}{%
    \begin{tabular}{lccccccc}\toprule
    Base Model   & Training  &  Inference & DocVQA & ChartQA & MathV   \\ \midrule
    \multirow{4}{*}{Qwen2-VL-2B~\cite{wang2024qwen2}}  &  \multirow{2}{*}{vanilla} & vanilla & 89.2 & 73.5 & 20.1	  \\ 
                                  &   & CoT &  84.9 & 55.5 & 13.8 \\
                                  &  \multirow{2}{*}{DocMark(w/o CoT)} & vanilla & 87.7 &  69.3 & 18.6\\   
                                  &  & CoT &  88.2 & 72.8  & 19.2 \\                           
                                  & DocMark(full) & CoT & 89.8 & 77.1 & 22.4 \\     \hline                             
\multirow{4}{*}{LLaVA-OneVision-7B~\cite{li2024llava}}  &  \multirow{2}{*}{vanilla} & vanilla &  	88.7 & 80.8	& 21.7  \\ 
                                  &  & CoT & 87.2 & 78.4 & 21.4 \\
                                  &  \multirow{2}{*}{DocMark(w/o CoT)} & vanilla & 87.0 & 79.1 & 21.5 \\   
                                  & & CoT & 88.2 & 80.4 & 22.0 \\                           
                                  & DocMark(full) & CoT & 90.1 & 83.4 & 23.2 \\                                  

	     \bottomrule
    \end{tabular}%
    }

\caption{\textbf{Comparison of model performance with and without CoT fine-tuning.}}
\label{tab:cot_ablation}
\end{table*}

\section{Performance Analysis}
\subsection{Qualitative Evaluation}

\begin{figure*}[htbp]
    \centering
    \includegraphics[width=0.88\linewidth]{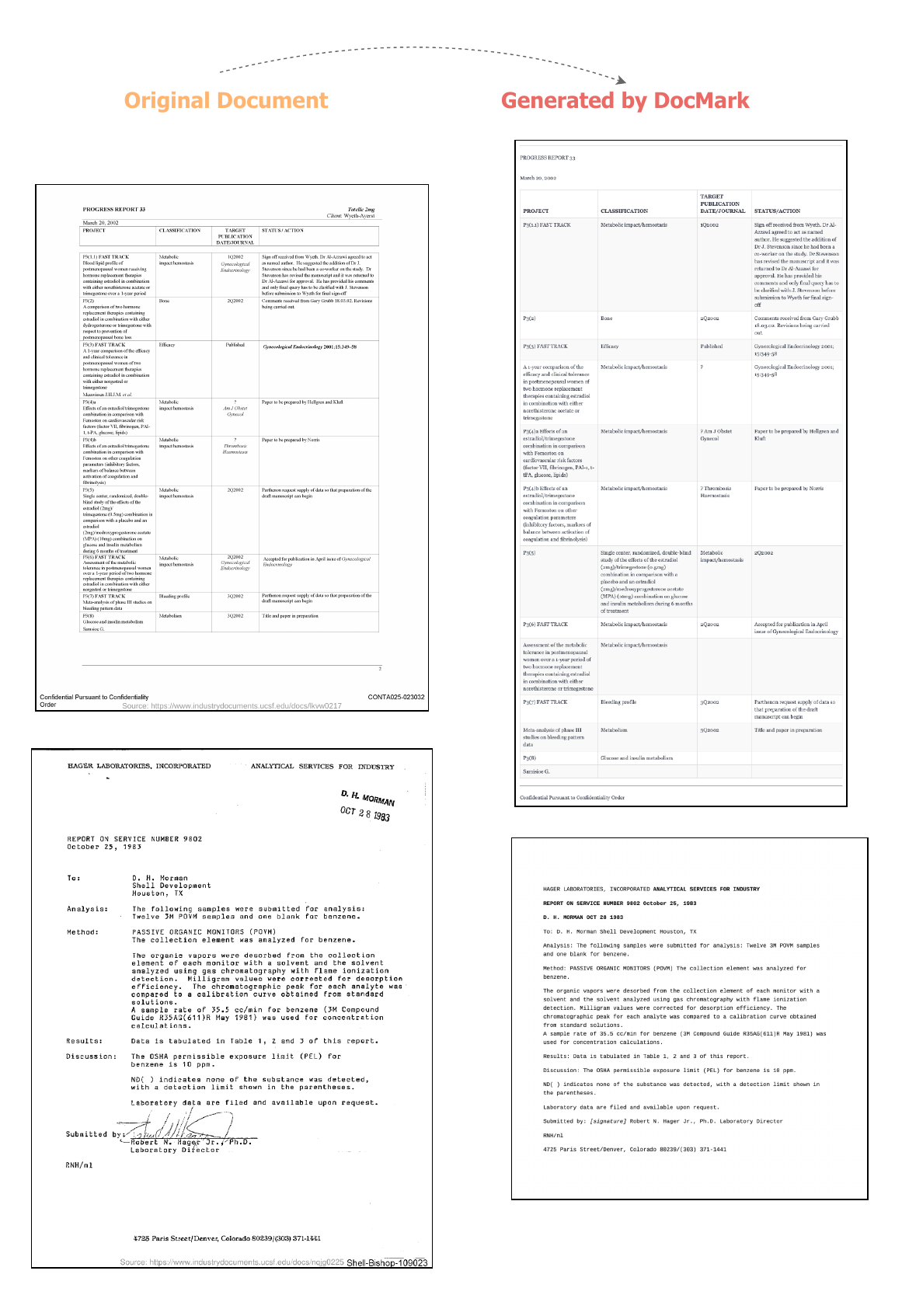}
    \vspace{-0.5em}
\caption{\textbf{Visualization on generated documents by DocMark.}}
    \label{fig:vis1}
    \vspace{-1.2em}
\end{figure*}

\begin{figure*}[htbp]
    \vspace{-2em}
    \centering
    \includegraphics[width=0.95\linewidth]{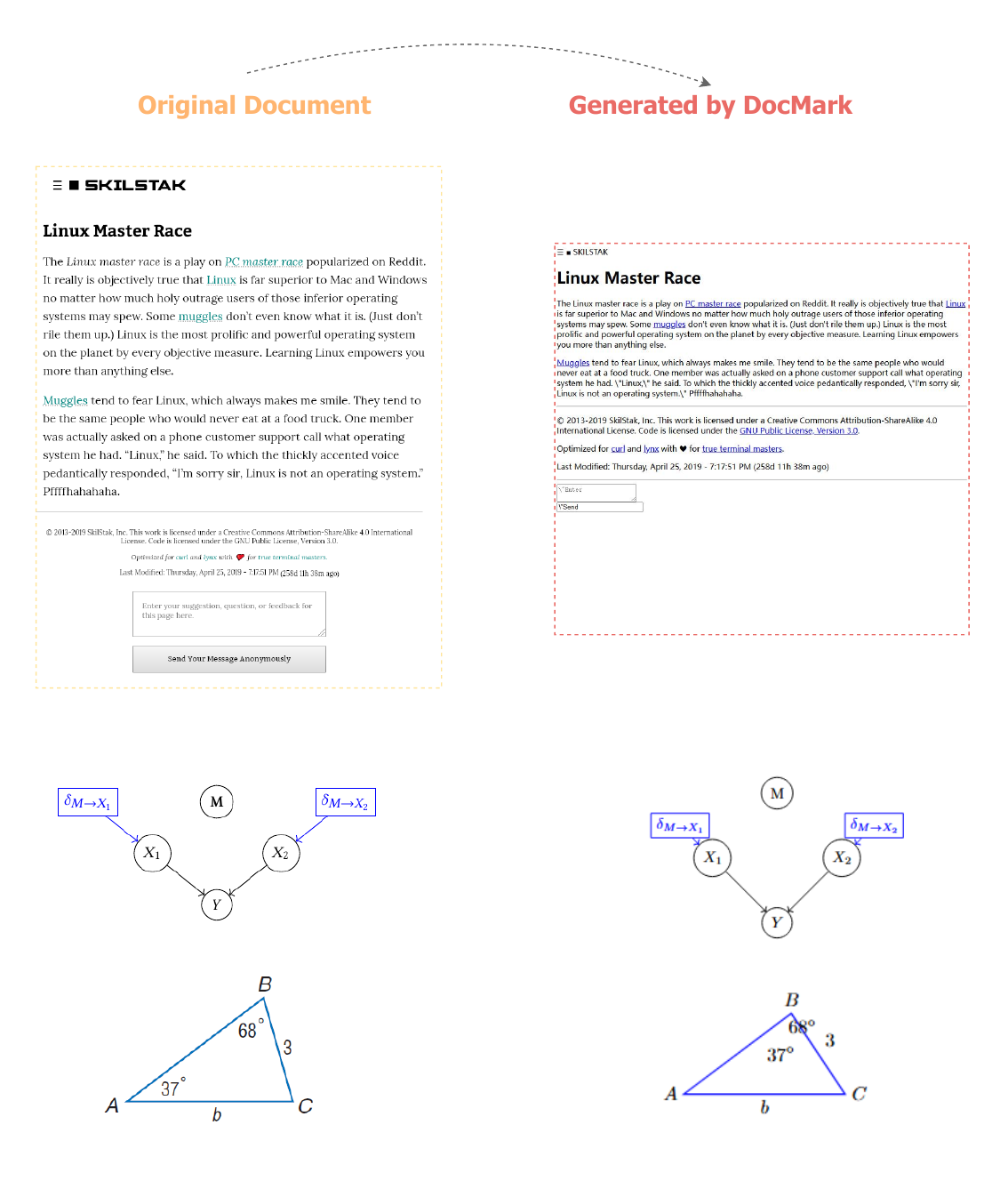}
    \vspace{-0.5em}
\caption{\textbf{Visualization on generated documents by DocMark.}}
    \label{fig:vis2}
    \vspace{8em}
\end{figure*}

For the qualitative evaluation, we offer more visualizations of the generated results by our DocMark, particularly on generating PDF documents, webpages, and scientific diagrams. As shown in \cref{fig:vis1} and \cref{fig:vis2}, our model is capable of maintaining the textual and layout information effectively. It should be noted that our method might not preserve the style information, such as font sizes and colors, very well. This is because our pre-training mainly concentrates on parsing the structured information within the documents. Learning the main textual and layout representations is sufficient for our document understanding tasks.

 \vspace{-2mm}
\subsection{Evaluating Accuracy with and without CoT Prediction}
We present a series of ablation experiments comparing: (1) vanilla models with direct CoT prompting, (2) models trained on DocMark-Pile and DocMark-Instruct with CoT fine-tuning removed, and (3) our full model.
As shown in \cref{tab:cot_ablation}, the vanilla models including Qwen2-VL and LLaVA-OneVision, despite being trained on extensive datasets, exhibit significant performance degradation compared to their unmodified counterparts, indicating a limited capacity for explicit reasoning.
The primary issue lies in these models' inability to effectively comprehend the document layout and derive accurate context from it. 
Notably, using CoT prompting with DocMark-Pile pre-trained models, even without CoT fine-tuning, surpasses the baseline performance.
Moreover, the vanilla models struggle to associate context with the question due to their inherent training methods that focuses on direct prediction. In contrast, our full approach, which incorporates end-to-end CoT fine-tuning, leads to superior performance.
This suggests that the enhancements in our original approach are primarily attributed to the effectiveness of the CoT component rather than solely introducing more data.

\begin{figure}[t]
    \vspace{0em}
    \centering
    \includegraphics[width=0.98\linewidth]{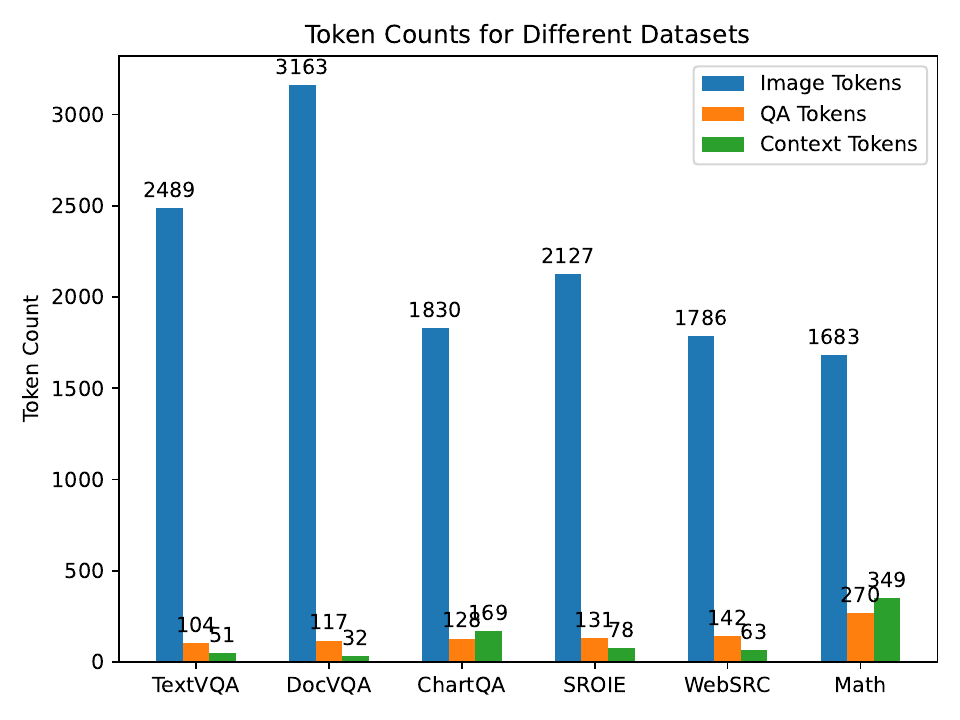}
    \vspace{0em}
\caption{\textbf{Token count comparison across various datasets.} We visualize the distribution of image tokens, context tokens, and original question-answer tokens on different datasets.}
    \label{fig:token}
    
\end{figure}

\subsection{Demonstrations of Adaptive Context Generation}
\begin{figure*}[t]
    \vspace{-2em}
    \centering
    \includegraphics[width=0.9\linewidth]{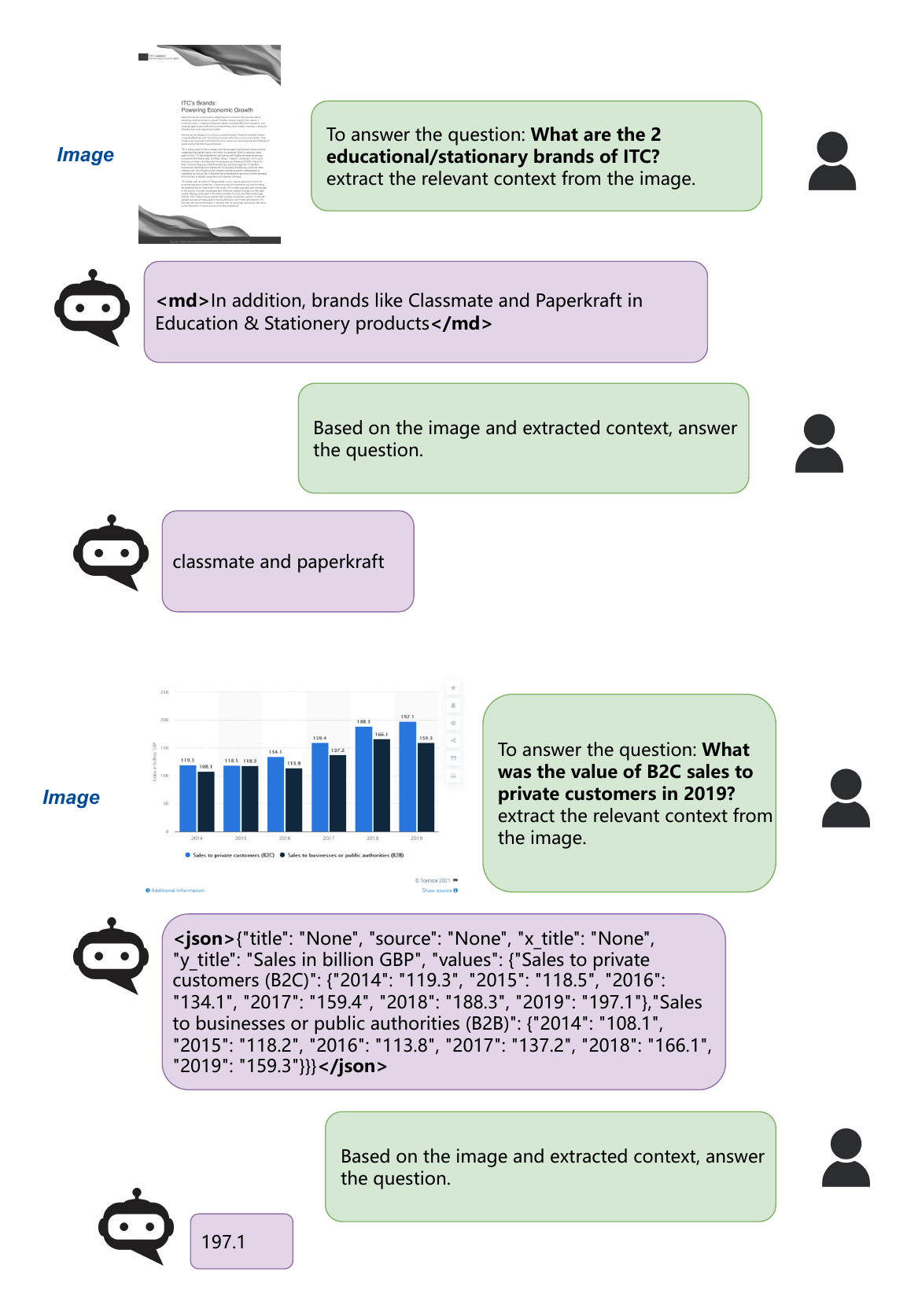}
    \vspace{-0.5em}
\caption{\textbf{Visualization on generated documents by DocMark.}}
    \label{fig:demo}
    \vspace{-1.2em}
\end{figure*}

To better showcase the adaptive generation capability of DocMark, we present additional demonstrations regarding downstream document understanding tasks in \cref{fig:demo}. This emphasizes our model's capacity to identify crucial information within documents and offer contextually-grounded answers, thereby enhancing the model's interpretability. By initially recognizing the document format, locating the area of interest, and extracting relevant structured representations, our model enhances both the accuracy of text recognition and the depth of understanding, which directly impacts the quality of the generated answers.

\subsection{Token Efficiency}
Since our adaptive generation pipeline incorporates additional auxiliary context tokens to derive final answers, we need to evaluate its impact on computational efficiency. 
To investigate this, we compare the number of image tokens, context tokens, and original question-answer tokens across several representative datasets. 
As demonstrated in \cref{fig:token}, image tokens account for the majority of the total token count due to the adopted dynamic resolution strategy. 
In contrast, context tokens contribute only a small number of tokens compared to both image tokens and original conversational tokens. This indicates that our method effectively provides highly condensed contextual information, alleviating the limitations of relying solely on image tokens and improving the overall understanding of the document with minimal computational cost.

\end{document}